\documentclass[10pt,twocolumn,letterpaper]{article}

\usepackage{wacv}
\usepackage{times}
\usepackage{epsfig}
\usepackage{graphicx}
\usepackage{amsmath}
\usepackage{amssymb}
\usepackage{algorithm}
\usepackage{algorithmic}

\usepackage{caption}
\usepackage{subcaption}
\usepackage{multirow}
\usepackage[T1]{fontenc}

\def\wacvPaperID{669} 

\wacvfinalcopy 

\ifwacvfinal
\def\assignedStartPage{9876} 
\fi
\ifwacvfinal
\usepackage[breaklinks=true,bookmarks=false]{hyperref}
\else
\usepackage[pagebackref=true,breaklinks=true,colorlinks,bookmarks=false]{hyperref}
\fi

\ifwacvfinal
\else
\fi

\begin{document}

\title{RNNP: A Robust Few-Shot Learning Approach}
\author{
Pratik Mazumder$^{\dagger}$ \hspace{2cm}Pravendra Singh$^{\dagger}$ \hspace{2cm}Vinay P. Namboodiri$^{\dagger \ast}$\\
$^{\dagger}$Department of Computer Science and Engineering, IIT Kanpur, India\\
$^{\ast}$University of Bath, United Kingdom \\
{\tt\small \{pratikm, psingh\}@iitk.ac.in, vpn22@bath.ac.uk}
}

\maketitle

\begin{abstract}
Learning from a few examples is an important practical aspect of training classifiers. Various works have examined this aspect quite well. However, all existing approaches assume that the few examples provided are always correctly labeled. This is a strong assumption, especially if one considers the current techniques for labeling using crowd-based labeling services. We address this issue by proposing a novel robust few-shot learning approach. Our method relies on generating robust prototypes from a set of few examples. Specifically, our method refines the class prototypes by producing hybrid features from the support examples of each class. The refined prototypes help to classify the query images better. Our method can replace the evaluation phase of any few-shot learning method that uses a nearest neighbor prototype-based evaluation procedure to make them robust. We evaluate our method on standard mini-ImageNet and tiered-ImageNet datasets. We perform experiments with various label corruption rates in the support examples of the few-shot classes. We obtain significant improvement over widely used few-shot learning methods that suffer significant performance degeneration in the presence of label noise. We finally provide extensive ablation experiments to validate our method.
\end{abstract}

\vspace{-10pt}
\section{Introduction}
\label{sec:intro}
Consider that you need to train a robot on a new concept of spectacles. You ideally would like to do so by providing a few examples. However, you make a mistake and provide an example of sunglasses as spectacles instead of using its original class. What will your robot do now? Will it retrieve your spectacles when you ask it to do so? Most probably not if your robot is trained using existing deep learning techniques. Through this paper, we address this issue and provide a method to perform robust prototype-based classification on a limited data setting.

Deep learning methods need vast amounts of annotated data for training, which can be very difficult and costly to obtain. Such models are unable to perform well for categories having very few training samples. On the other hand, humans can quickly learn new categories by simply looking at a few samples and can perform very well in identifying them. We need to reduce this gap between deep learning and humans. Few-shot learning algorithms address this shortcoming of deep learning. Few-shot learning involves training networks in such a way that they can achieve good results for classes that have very few training data (few-shot classes). Researchers have proposed several methods to deal with this problem \cite{snell2017prototypical,finn2017model,rusu2018meta,oreshkin2018tadam,triantafillou2017few,hsu2018unsupervised,wei2019piecewise}. However, it is possible that these few-shot classes contain some support examples with corrupted labels. This makes the few-shot learning problem even more challenging. In this paper, we deal with the problem of few-shot learning, where the few-shot classes contain support examples with corrupted labels. In the few-shot learning setting, the network receives one episode at a time, and each episode contains a set of support examples and a set of query examples. Each episode contains a few classes and has very few support examples per class. The query examples also belong to one of the classes present in the episode. In this setting, the task of the network is to use the limited support data present in an episode to classify the query data points in the episode.

During the base class (defined in Sec.~\ref{sec:setting}) training phase, the model trains on episodes containing classes from the base class set only. During the novel class (defined in Sec.~\ref{sec:setting}) evaluation phase (testing), the model receives episodes from novel/few-shot classes (not part of the base class set), and it has to successfully classify the query examples of the novel class episode using the support examples in that episode only. Since the number of support examples per class is very low, standard classification training makes the network overfit to the limited training data in the episode and fail miserably in classifying the query data. A large number of few-shot learning methods perform few-shot testing by applying the nearest neighbor prototype-based testing approach (NNP) \cite{snell2017prototypical,oreshkin2018tadam,ye2018learning,xing2019adaptive,wang2018low,li2019finding}, i.e., for each class, they learn a prototype, and for each query example, they calculate the distance of the query feature from the prototypes. They predict the class prototype closest to the query example in the feature space as the output class.

In the above setting, during the novel class evaluation phase, if some of the support examples of the novel/few-shot classes have corrupted labels, then the few-shot learning algorithms will suffer from significant performance degradation. We propose a novel method to address the loss in performance (due to corrupted support examples) of few-shot learning methods that use the NNP testing approach. This work focuses specifically on the novel class evaluation phase (testing) of few-shot learning algorithms. Several methods deal with corrupted labels in the full-shot settings (with a huge number of samples per class) \cite{sanderson2014class,jiang2017mentornet,malach2017decoupling,han2018co,yu2019does}. However, these methods are training-based and do not perform well when trained on the few support examples in the novel class episode due to overfitting, as shown in Sec. \ref{sec:jocor}. Therefore, there is an urge to make few-shot learning robust to label corruption.

\begin{figure*}
\begin{tabular}{ccc}
\centering
\frame{\includegraphics[width=0.31\textwidth]{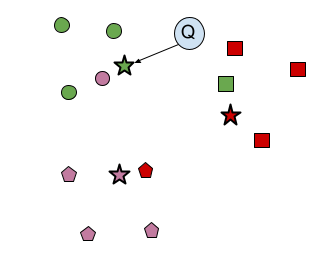}}&
\frame{\includegraphics[width=0.31\textwidth]{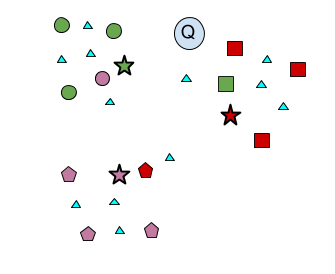}}&
\frame{\includegraphics[width=0.31\textwidth]{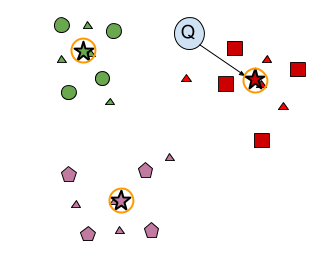}}\\
(a)&(b)&(c)
\end{tabular}
\vspace{-10pt}
\caption{(a) Green, red, and magenta refer to support examples of 3 classes, say A, B, and C, respectively. Let us assume that due to label corruption, 1 support example of each class is corrupted (marked with the color of a different class). Stars represent the prototype for each class (marked with the color of the corresponding class) computed by averaging the support features. Standard nearest neighbor prototype-based classification (NNP) results in wrongly predicting class A for query data point Q. (b, c) Triangles represent the unlabeled hybrid features produced from the support example features. RNNP produces unlabeled hybrid features from the support examples and performs clustering on the hybrid features, support features, and the query feature. It updates the cluster centers and rectifies the labels of the support examples. The newly computed class prototypes (depicted as stars marked with the color of the corresponding class and surrounded by an orange circle) result in correctly predicting class B for the query data point Q. The arrow points to the class prototype closest to Q.}
\label{fig:rnnp}
\vspace{-10pt}
\end{figure*}

The support examples with corrupted labels (in a novel class episode) corrupt the prototypes computed for each class and harm the performance. Our method refines the prototypes in order to reduce the level of label corruption in them. It produces hybrid features from the features of the support examples extracted by the network and performs a soft k-means clustering on the support features and hybrid features in order to refine the prototypes. Since the corrupt support examples are relatively closer to the other support examples of their actual class, the clustering procedure helps correct the labels for many such support examples. This, in turn, helps reduce the influence of the corrupt support examples in the class prototypes of the wrong class. Consequently, the refined class prototypes perform better in classifying the query data points of that episode. We call our method the Robust Nearest Neighbor Prototype-based testing approach (RNNP) (depicted in Fig.~\ref{fig:rnnp}). Our method can replace the testing process of any few-shot learning algorithm that uses the nearest neighbor prototype-based evaluation. Our method does not make use of any external data or additional labels of any form. In order to demonstrate the efficacy of our method, we replace the testing process of a state-of-the-art few-shot learning method FEAT~\cite{ye2018learning} and a very popular few-shot learning method prototypical network \cite{snell2017prototypical}. Our experiments show that our testing strategy works for multiple datasets and few-shot learning methods. 

To the best of our knowledge, this is the first work that tries to make the testing strategy of few-shot image classification algorithms robust to any label corruption in the support examples. Our major contributions are as follows:

\begin{itemize}
    \item We propose a novel approach called Robust Nearest Neighbor Prototype-based testing (RNNP) that makes the few-shot evaluation process robust to noisy/corrupted labels in the novel class episodes.
    \item We experimentally show that our method significantly improves the performance of few-shot learning methods that use a nearest neighbor prototype-based testing strategy, in the presence of noisy labels.
    \item We validate the different components of our method using extensive ablation studies.
\end{itemize}

\vspace{-10pt}
\section{Related Works}
\subsection{Few-Shot Classification}
 Researchers have proposed several approaches to deal with the problem of few-shot classification.
 Siamese Neural Network \cite{koch2015siamese} trained models to learn to discriminate between pairs of images. This training strategy helped the network to even discriminate between images from previously unseen classes at test time. Prototypical Network \cite{snell2017prototypical} produces class prototypes by averaging the features of support examples of that class and then uses the nearest neighbor approach to find the nearest prototype to the query feature in order to predict its class. 
 Model Agnostic Meta Learning \cite{finn2017model} method trains the network to quickly adapt to new data by taking an average gradient step that benefits multiple non-similar tasks at once.
 RelationNet \cite{sung2018learning} trains the network to generate a similarity score between a query image and support examples of a class. 
 
 PFA \cite{qiao2018few} tries to predict the parameters for adapting a network to the new class by using the activations for new images. TADAM \cite{oreshkin2018tadam} uses task-based embedding as attention to the network to produce better features for the images in the episode. 
 MetaOptNet \cite{lee2019meta} trains the network to learn features that work well with linear classifiers when used for few-shot classification. CTM \cite{li2019finding} trains a module to select important dimensions in the features that help improve the classification performance. 
 Simple Shot \cite{wang2019simpleshot} optimizes the representations to perform better on the nearest neighbor classifier.

 FEAT - Few-shot Embedding Adaptation with Transformer \cite{ye2018learning} uses multi-headed attention to adapt the class prototypes to draw them away from each other. This helps the nearest neighbor classification further. PARN \cite{wu2019parn} trains a position-aware relation network (PARN) that tries to make the relation module invariant to changes in the spatial position of the object.
 MetaMG \cite{xie2019meta} trains a network to perform fast few-shot incremental learning

\subsection{Learning with Noisy Labels}\label{sec:relnoise}
Researchers have proposed various methods to helps the network learn better in the presence of noisy labels. Some methods focus on estimating the latent noisy transition matrix \cite{liu2015classification,menon2015learning,sanderson2014class,patrini2017making}. Co-teaching \cite{han2018co}, involves training two networks simultaneously, and each network provides samples with possibly clean labels from each mini-batch to the other network for training.

Methods like Decoupling \cite{malach2017decoupling} use a ``Disagreement" strategy of training, where the network updation is based on the level of disagreement between two different networks on the same data point. Co-teaching+ \cite{yu2019does} combines the ``Disagreement" strategy with Co-teaching to improve the robustness of the model. However, there is no guarantee that the examples chosen for the ``Disagreement" strategy have the correct labels. Another approach of tackling this problem is to utilize regularization from peer networks to improve the generalization ability of networks by encouraging agreement between them. However, these methods still suffer from the memorization of the noisy labels \cite{zhang2016understanding}. JoCoR (Joint Training with Co-Regularization) \cite{Wei_2020} trains two networks with a conventional supervised loss and a co-regularization loss. It also employs an example selection strategy, which chooses examples with a lower total loss for training the networks. This minimizes the flow of erroneous information.

Our few-shot setting has corrupt labels only in the novel class episodes, which have very few samples. These methods require a large number of labeled data for training, and therefore, they overfit to the scant data in the few-shot setting. We experimentally show in Sec. \ref{sec:jocor} that such methods do not perform well and are not suitable for the few-shot setting.

\vspace{-10pt}
\section{Proposed Method}
\label{sec:method}
\subsection{Problem Setting}\label{sec:setting}
In the few-shot learning setting, there are two sets of classes of images, i.e., the base class set and the novel class set. The unique nature of this setting is that these two sets of classes do not share any common classes. The base classes $C_{base}$ have many labeled data points, but the novel/few-shot classes $C_{few}$ only have a few labeled examples. 

An episodic formulation is generally followed in the few-shot learning setting. In this setting, the network receives 1 episode at a time and each episode contains a set of support examples and a set of query examples. In an episode, there are N classes and each class has K support examples per class i.e. $S=\{ (x_1,y_1),(x_2,y_2). . .(x_{K \times N},y_{K \times N})\}$. The query examples also belong to one of the N classes i.e. $Q=\{ (x_1^{\ast},y_1^{\ast}),(x_2^{\ast},y_2^{\ast}). . .(x_{q}^{\ast},y_{q}^{\ast}\}$. Here $x_{i}$ are images and $y_{i} \in \{ 1 .. N \}$ are labels. Such an episode is referred to as a K-shot N-way episode. The number of support examples per class is very low (e.g. K=1,5,10).

There are two phases in this setting: the base class training phase and the novel class evaluation (testing) phase. During the base class training phase, the network trains on episodes of the base classes. Each such episode contains a few classes from the base class set. The network trains on the support examples per class $(S)$ present in each episode. During this phase, the network can carry over the knowledge gained from one episode to the next.

During the novel class evaluation phase, the network receives episodes of the novel classes. The network has to use the few support examples per class $(S)$ present in an episode to classify the query data points $(Q)$ in the episode. During this phase, the network cannot carry over the knowledge gained from one episode to the next. In the novel class episodes, some of the support examples labels are incorrect. This type of corruption makes the few-shot classification of the novel classes challenging. The distribution of label corruption is uniform across the classes in each episode. The objective of the network is to deal with this corruption and classify query examples in each novel class episode.

\subsection{Method Overview}
In the few-shot learning problem setting with label corruption (in novel class episodes), since there are a limited number of support examples per class and some support examples may also be corrupted (belonging to a different class), the computed class prototypes are not correct. We propose a novel method (RNNP) to improve the robustness of the few-shot learning method when used for few-shot testing on episodes that contain support examples with corrupted labels (Fig.~\ref{fig:rnnp}). Since RNNP is a testing method, it can replace the testing process of any few-shot learning method that utilizes a nearest neighbor prototype-based (NNP) few-shot testing mechanism. 

Let us assume that we have a base network $B$ already trained on the episodes from the base classes. During the novel class evaluation phase (testing), given a novel class episode, we first extract features for the support images and query images using the trained base network ($B$).
\begin{equation}\label{eq:extract}
        z_i = B(x_i)
\end{equation}
where $x_i$ refers to the support or query image, and $z_i$ is the extracted feature of that image.

Next, we calculate the prototypes of each class using the support image features of that class. Many methods use simple averaging to get the mean prototype per class \cite{snell2017prototypical}. Some methods adjust this mean prototype to further help in classifying the query examples \cite{ye2018learning}. 
\begin{equation}\label{eq:proto}
    p_n = \sum\limits_{i=1}^K \frac{z_i^n}{K}
\end{equation}
where $p_n$ is the prototype for the $n^{th}$ class and $z_1^n,z_2^n, . . .z_K^n$ are the features of the support examples of the $n^{th}$ class.

The computation process of the prototypes unknowingly utilizes the support examples with corrupted labels too. Therefore, the class prototypes are not fully correct and need refinement. In order to refine them, we first create unlabeled hybrid features by combining the support images features from the same class using proportion hyperparameter $\alpha$. For each support image feature, we generate $\beta$ unlabeled hybrid features.
\begin{equation}
    z_{u}=\alpha * z_i^n + (1- \alpha) * z_j^n
\end{equation}
where $z_{u}$ is the generated unlabeled hybrid feature, $z_i^n, z_j^n$ are features of support examples of class $n \in C_{few}$ and $i \neq j$. $\alpha \in (0,1)$  is the proportion hyper-parameter.

Next, for each query image, we perform soft k-means clustering on the combined set of the support image features, the hybrid features, and the corresponding query image feature. We use the class prototypes computed in Eq. \ref{eq:proto} as the initial cluster centers for the clustering algorithm. Using the current cluster centers, we assign soft labels to the support and hybrid features and the single query feature. We update the cluster centers using these soft labels. We repeat this process a few times and use the final cluster centers as the refined class prototypes. Since the corrupt support features are relatively closer to the features of the support examples of their actual classes, the clustering process can correctly assign many of them to their actual classes. This process serves two purposes. Firstly, it reduces the influence of the corrupted labels on the class prototypes. Secondly, this process also reduces the problem of overfitting in the prototypes because we compute the refined prototypes using both hybrid and support features.

After obtaining the refined prototypes, we find the distance from each of refined class prototypes in the feature space to the query image. Next, we apply a softmax function over the negative values of the distances so that refined class prototypes at smaller distances from the query example get higher softmax probabilities.
\begin{equation}\label{eq:distance}
    W^i_q = \frac{exp(-d(z_q,p^{*}_i))}{\Sigma_{j=1}^N exp(-d(z_q,p^{*}_j))}
\end{equation}
where $W^i_q$ is the class probability for $q^{th}$ query example for the $i^{th}$ class, d(.,.) is the distance metric, $p^{*}_i$ is the refined class prototype for the $i^{th}$ class, $z_q$ is the feature for the $q^{th}$ query example. $N$ is the total number of classes in the episode. 

The class for which the above probability is highest, is predicted as the output class for that query image.
\begin{equation}\label{eq:predict}
        \hat{y}_q = argmax_n W^n_q
\end{equation}
where $n\in\{1,2,...,N\}$ refers to the $n^{th}$ class, $\hat{y}_q$ refers to the predicted class for the $q^{th}$ query example.

\subsection{Different from Transductive and Semi-Supervised FSL}
Our method should not be confused with transductive few-shot learning. Transductive few-shot learning makes use of all the query points simultaneously to help in classifying themselves by making use of the structure of this bulk data. However, our setting is the standard few-shot setting, and we cannot use the query images to help classify other query images. Instead, we utilize hybrid features to help in classifying the query images better.

Our method also does not fall under the semi-supervised few-shot learning setting. Semi-supervised few-shot learning uses unlabeled real data from the same dataset, either belonging to the classes in that episode or to other classes in the dataset. These data are used for training and also for testing. However, our method does not use extra real data and only uses an interpolation of support features of the same class.This makes our method very efficient. We consider them as unlabeled data points and use them in our clustering algorithm only during testing.

\section{Experiments}
\subsection{DataSets}
We perform few-shot classification experiments on 2 few-shot classification datasets, namely, mini-ImageNet \cite{vinyals2016matching} and tiered-ImageNet \cite{ren2018meta}. 

mini-ImageNet \cite{vinyals2016matching} is the most popular few-shot learning dataset. It is derived from the ImageNet dataset \cite{russakovsky2015imagenet}. It has 100 classes and 600 images of size $84\times 84$ pixels for each class. The classes consist of 64 train classes, 16 validation classes, and 20 test classes.

tiered-ImageNet \cite{ren2018meta} is larger subset of the ImageNet \cite{russakovsky2015imagenet} dataset. It has 608 classes divided into 351 train, 97 validation, and 60 test classes. It groups similar classes into higher-level classes. In tiered-ImageNet, the training classes differ significantly from the testing classes as compared to mini-ImageNet.

\begin{table*}[t]
\renewcommand\arraystretch{1.1}
\footnotesize
\centering
\caption{Performance of FEAT and ProtoNet with NNP/RNNP on the mini-ImageNet dataset over 5/10-shot 5-way episodes with 0\%, 20\%, 40\% label corruption.}
\vspace{-5pt}
\addtolength{\tabcolsep}{3pt}
\begin{tabular}{c|l|c|c|c|c}
\hline
Corruption & Testing & \multicolumn{2}{c|}{FEAT (CVPR'20)} & \multicolumn{2}{c}{ProtoNet (NIPS'17)}  \\
  \cline{3-6} 
Rate & Method & 5-shot & 10-shot & 5-shot & 10-shot  \\
  \hline
0\% & NNP & 82.05 $\pm$ 0.14 \% & 85.03 $\pm$ 0.44\%  & 80.53 $\pm$ 0.14 \% & 83.61 $\pm$ 0.45\% \\
0\% & RNNP (Ours) & \textbf{82.76} $\pm$ 0.42\% & \textbf{85.52} $\pm$ 0.46\% & \textbf{81.62} $\pm$ 0.45\% & \textbf{83.63} $\pm$ 0.47\%\\
\hline
20\% & NNP & 76.67 $\pm$ 0.44\%  & 82.23 $\pm$ 0.47\%  & 75.57 $\pm$ 0.45\% &  80.91 $\pm$ 0.43\% \\
20\% & RNNP (Ours) & \textbf{78.98} $\pm$ 0.42\% &  \textbf{83.68} $\pm$ 0.43\%  & \textbf{76.72} $\pm$ 0.42\% & \textbf{81.69} $\pm$ 0.46\% \\
\hline
40\% & NNP & 65.07 $\pm$ 0.58\% & 74.37 $\pm$ 0.60\% & 64.63 $\pm$ 0.55\% & 73.27 $\pm$ 0.57\% \\
40\% & RNNP (Ours) & \textbf{70.97} $\pm$ 0.57\% & \textbf{79.72} $\pm$ 0.54\% & \textbf{69.02} $\pm$ 0.60\% & \textbf{77.52} $\pm$ 0.58\%  \\
\hline

\hline

\end{tabular}
\label{tab:corruptmini}
\vspace{-5pt}
\end{table*}

\begin{table*}[t]
\renewcommand\arraystretch{1.1}
\footnotesize
\centering
\caption{Performance of FEAT and ProtoNet with NNP/RNNP on the tiered-ImageNet dataset over 5/10-shot 5-way episodes with 0\%, 20\%, 40\% label corruption.}
\vspace{-5pt}
\addtolength{\tabcolsep}{3pt}
\begin{tabular}{c|l|c|c|c|c}
\hline
Corruption & Testing & \multicolumn{2}{c|}{FEAT (CVPR'20)} & \multicolumn{2}{c}{ProtoNet (NIPS'17)}  \\
  \cline{3-6} 
Rate & Method & 5-shot & 10-shot & 5-shot & 10-shot  \\
  \hline
0\% & NNP & 84.79 $\pm$ 0.16\% & 87.72 $\pm$ 0.46\%  & 84.03 $\pm$ 0.16\% & 87.25 $\pm$ 0.45\% \\
0\% & RNNP (Ours)& \textbf{85.57} $\pm$ 0.45\%  & \textbf{88.35} $\pm$ 0.43\% & \textbf{84.77} $\pm$ 0.45\% & \textbf{87.25} $\pm$ 0.47\% \\
\hline
20\% & NNP & 80.07 $\pm$ 0.43\%  & 85.35 $\pm$ 0.45\%  & 80.23 $\pm$ 0.44\% &  84.95 $\pm$ 0.42\% \\
20\% & RNNP (Ours) & \textbf{81.69} $\pm$ 0.44\% &  \textbf{86.52} $\pm$ 0.43\% & \textbf{81.25} $\pm$ 0.46\% & \textbf{85.79} $\pm$ 0.45\% \\
\hline
40\% & NNP & 69.57 $\pm$ 0.58\% & 79.17 $\pm$ 0.55\% & 69.08 $\pm$ 0.54\% & 78.61 $\pm$ 0.59\%\\
40\% & RNNP (Ours) & \textbf{73.04} $\pm$ 0.59\% & \textbf{84.55} $\pm$ 0.54\% & \textbf{72.35} $\pm$ 0.58\% & \textbf{82.54} $\pm$ 0.55\%  \\
\hline

\hline

\end{tabular}
\label{tab:corrupttiered}
\vspace{-15pt}
\end{table*}

\subsection{Implementation Details}
Since the mini-ImageNet and tiered-ImageNet datasets do not contain corrupted labels, we manually corrupt the support image labels for the novel classes during the novel class evaluation phase. We perform experiments with 0\%, 20\%, and 40\% label corruption rates in the support examples of the novel classes. 20\% label corruption rate refers to the setting where 20\% of the support examples in an episode have incorrect labels. The distribution of label corruption is uniform across all the classes in each episode. For each novel class episode, we assume that the support image with corrupted labels belongs to another class from the same episode. We use this assumption so that the episodic formulation holds, i.e., all the examples in an episode are from the classes in the episode. For fairness, we perform all the experiments in this setting. We perform experiments in this corrupted label setting for 5-shot 5-way and 10-shot 5-way episodes.

We replace the testing process of two few-shot learning methods (ProtoNet and FEAT) with our proposed method, RNNP, and perform few-shot testing experiments on the trained models. We do not retrain the models. The first method is the prototypical network (ProtoNet)\cite{snell2017prototypical}, which is a very popular few-shot learning method. It trains and tests the network on episodes and predicts the output class using the nearest neighbor prototype-based (NNP) approach to find the closest class prototype to the query feature. We also perform experiments on the recent state-of-the-art few-shot learning method FEAT - (Few-shot Embedding Adaptation with Transformer) \cite{ye2018learning}. FEAT adapts the class prototypes using a multi-headed self-attention based transformer to increase the gap between them. It also uses NNP for testing. We use ResNet-12 \cite{ye2018learning} as the backbone architecture.

We follow the testing protocol that we have defined earlier. Whenever we calculate/re-calculate the class prototypes (cluster centers), we use the mechanism proposed by the respective method (FEAT or ProtoNet) for each case. For the prototypical network (ProtoNet), we compute the class prototypes by averaging the support image features of the corresponding class. For FEAT, we compute the class prototypes by averaging the support image features of the corresponding classes and adapting them using a trained transformer.

We can choose the proportion hyper-parameter $\alpha$ between 0 and 1. We use $\alpha=0.8$ for our experiments and validate this choice through ablation experiments. In the 5-shot episodes, for each support example, we have 4 other support examples of the same class. Similarly, in the 10-shot episodes, for each support example, we have 9 other support examples of the same class.  For each support example in the 5-shot episodes, we generate $\beta=4$ hybrid features using the 4 other support examples of the same class. Similarly, for each support example in the 10-shot episodes, we generate $\beta=9$ hybrid features. For our experiments, we repeat the clustering and label assignment process for three times. We validate these choices using ablation experiments.
We evaluate both the models with and without RNNP. We evaluate the models for 1000 episodes and report the best out of 5 runs for each method. The above-mentioned methods (FEAT, ProtoNet) with the traditional testing protocol form our baseline for those methods. 

Since there are no previous works that make the few-shot evaluation procedure robust to noisy support examples of novel classes, we compare our method with two popular full-shot noisy label training methods in our setting (Sec.~\ref{sec:jocor}).

\subsection{mini-ImageNet Results}
Table \ref{tab:corruptmini} reports the results for few-shot classification on 5-shot 5-way episodes of mini-ImageNet with various levels of corruption. The results indicate that for both FEAT and ProtoNet, the performance drops significantly with the increase in the level of corruption of the support image labels when using the standard nearest neighbor prototype-based testing process (NNP). For 20\% label corruption, the performance of FEAT drops by absolute margins of 5.38\% and 2.8\% for 5-shot and 10-shot 5-way episodes, respectively. Similarly, for 20\% label corruption, the performance of ProtoNet drops by absolute margins of 4.96\% and 2.7\% for 5-shot and 10-shot 5-way episodes, respectively. When our method RNNP replaces the testing process of FEAT and ProtoNet, their performances improve significantly. Specifically, for FEAT, using RNNP improves the performance in the 20\% label corruption setting by absolute margins of 2.31\% and 1.45\% for 5-shot and 10-shot 5-way episodes, respectively. The performance in the 40\% label corruption setting improves by absolute margins of 5.9\% and 5.35\% for 5-shot and 10-shot 5-way episodes, respectively. As the level of corruption increases, the performance improvement due to our method also increases. Even when the label corruption rate is 0\%, our method does not have any side effect on the performance. In fact, it increases the performance in some cases.

\begin{figure}[t]
\includegraphics[width=0.23\textwidth]{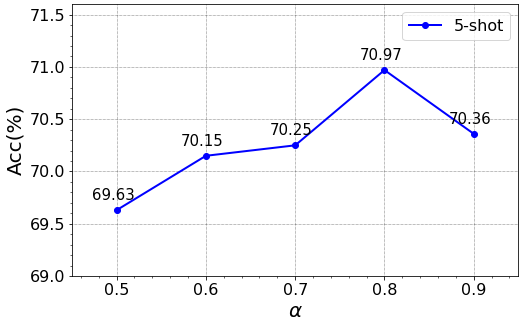}
\includegraphics[width=0.23\textwidth]{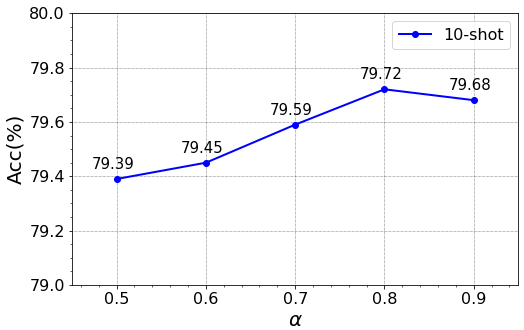}
\vspace{-10pt}
\caption{Performance of FEAT with RNNP on the mini-ImageNet dataset over 5/10-shot 5-way episodes with 40\% label corruption for different values of $\alpha$.}
\label{fig:alpha}
\vspace{-10pt}
\end{figure}

\subsection{tiered-ImageNet Results}

Table \ref{tab:corrupttiered}, reports the results for few-shot classification on 5-shot 5-way episodes of tiered-ImageNet with various levels of corruption. 
For 20\% label corruption, the performance of FEAT drops by absolute margins of 4.72\% and 2.37\% for 5-shot and 10-shot 5-way episodes, respectively. Similarly, for 20\% label corruption, the performance of ProtoNet drops by absolute margins of 3.8\% and 2.3\% for 5-shot and 10-shot 5-way episodes, respectively. When our method RNNP replaces the testing process of FEAT and ProtoNet, their performances improve significantly. Specifically, for FEAT, the performance in the 40\% label corruption setting improves by absolute margins of 3.47\% and 5.38\% for 5-shot and 10-shot 5-way episodes, respectively.

\section{Ablation}

\subsection{Value of $\alpha$}

Fig.~\ref{fig:alpha} depicts the performance of FEAT with our proposed RNNP for different values of $\alpha$. Since we combine the support features using $\alpha$ and $1-\alpha$, therefore, we perform ablation experiments with $\alpha$ values from $0.5$ to $0.9$. We observe the best results for $\alpha=0.8$ for both 5-shot 5-way and 10-shot 5-way episodes. We use this value of $\alpha$ for all our experiments.

\subsection{Value of $\beta$}

\begin{figure}[t]
\includegraphics[width=0.23\textwidth]{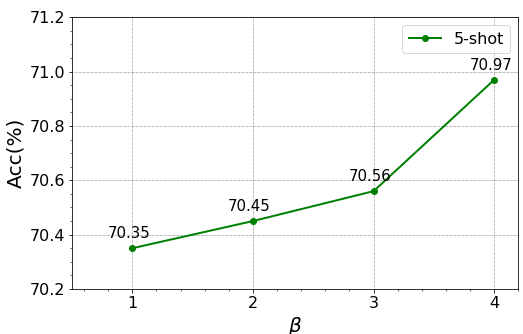}
\includegraphics[width=0.23\textwidth]{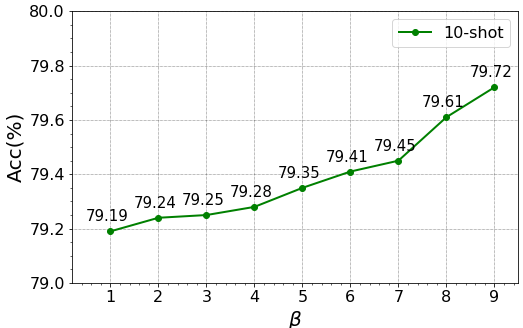}
\vspace{-10pt}
\caption{Performance of FEAT with RNNP on the mini-ImageNet dataset over 5/10-shot 5-way episodes with 40\% label corruption for different values of $\beta$.}
\label{fig:beta}
\vspace{-10pt}
\end{figure}

Fig.~\ref{fig:beta} depicts the performance of FEAT with our proposed RNNP for different values of $\beta$. We observe the best results for $\beta=4$ and $\beta=9$ in the 5-shot and 10-shot 5-way episodes, respectively, which are also the maximum possible value for $\beta$ in the respective cases.

\subsection{Number of Iterations}

\begin{figure}[t]
\includegraphics[width=0.23\textwidth]{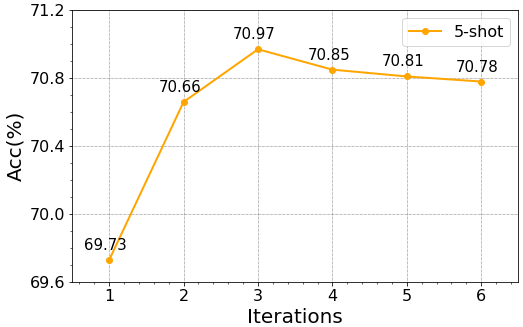}
\includegraphics[width=0.23\textwidth]{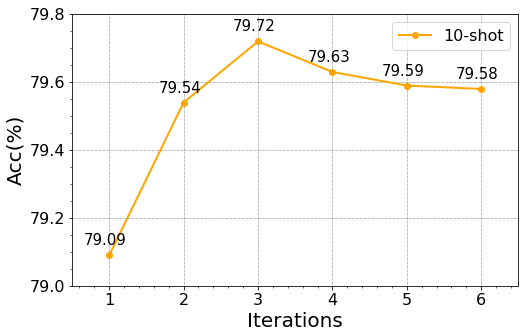}
\vspace{-5pt}
\caption{Performance of FEAT with RNNP on the mini-ImageNet dataset over 5/10-shot 5-way episodes with 40\% label corruption for different number of iterations of the clustering process.}
\label{fig:count}
\end{figure}

Fig.~\ref{fig:count} depicts the performance of FEAT with our proposed RNNP for different numbers of iterations of the clustering process used in our method. We observe the best results for 3 iterations of our process for both 5-shot and 10-shot settings. 

\subsection{Labeled vs. Unlabeled Hybrid Features}

In our method, we keep the hybrid features as unlabeled. The hybrid features may not necessarily belong to the class of the parent features since classification boundaries are usually non-linear, and in such cases, points that lie between 2 features from the same class may not necessarily belong to that class. Further, because of the presence of noisy labels, the hybrid features may have been created by combining features of two different classes. To validate this point, we performed an experiment using the FEAT w/ RNNP model with 40\% label corruption, assuming that the hybrid features belonged to the class of the parent features and directly updated the class prototypes without using soft k-means. The resulting model performed similarly to the FEAT model without RNNP. This is why we keep the hybrid features as unlabeled.

\subsection{Hard Clustering vs. Soft Clustering}

\begin{table}[t]
\renewcommand\arraystretch{1.1}
\footnotesize
\centering
\vspace{-5pt}
\caption{Performance of FEAT with RNNP on the mini-ImageNet dataset over 5/10-shot 5-way episodes with 40\% label corruption using hard/soft k-means.}
\addtolength{\tabcolsep}{3pt}
\begin{tabular}{l|c|c}
\hline
K-means & \multicolumn{2}{c}{FEAT}  \\
  \cline{2-3} 
 & 5-shot & 10-shot  \\
  \hline

Hard & 70.63 $\pm$ 0.55\% & 79.23 $\pm$ 0.55\% \\
Soft & \textbf{70.97} $\pm$ 0.57\% & \textbf{79.72} $\pm$ 0.54\% \\
\hline

\hline

\end{tabular}
\label{tab:softhard}
\vspace{-10pt}
\end{table}

In our experiments, we use a soft k-means algorithm for performing clustering on the features. Hard k-means clustering assigns points to a single cluster, whereas in soft k-means clustering, the points do not fully belong to any one cluster. We perform an ablation to check whether hard k-means clustering performs better than soft k-means clustering when used in our method. From Table \ref{tab:softhard}, we can see that RNNP based on soft k-means clustering performs better than RNNP based on hard k-means clustering. Soft k-means clustering allows the features to refine multiple clusters (based on their distance from the cluster centers), and this helps it to perform better than the hard k-means algorithm in this case.

\subsection{Hybrid Features from Different Classes}
\begin{table}[t]
\footnotesize
\centering
\caption{Performance of FEAT with RNNP on the mini-ImageNet dataset over 5/10-shot 5-way episodes with 40\% label corruption by creating hybrid features from the same class or from different classes.}
\begin{tabular}{c|c|c}
\hline
Same/Different class & 5-shot & 10-shot   \\
  \hline

Different & 70.63 $\pm$ 0.60 \% & 79.21 $\pm$ 0.56\%  \\
Same & \textbf{70.97} $\pm$ 0.57\% & \textbf{79.72} $\pm$ 0.54\% \\
\hline

\end{tabular}
\label{tab:ablsamedif}
\end{table}

Table \ref{tab:ablsamedif} depicts the performance of FEAT with our proposed RNNP using hybrid features produced by combining support features from the same or different classes. The results indicate that we achieve better results when we use support features from the same class rather than from different classes for 5-shot and 10-shot episodes.

\subsection{Hybrid Features vs. Noise}

\begin{table}[t]
\footnotesize
\centering
\vspace{-10pt}
\caption{Performance of FEAT with RNNP on the mini-ImageNet dataset over 5/10-shot 5-way episodes with 40\% label corruption by using noise, hybrid features, features of new images generated using Mixup or CutMix data augmentation.}
\begin{tabular}{l|c|c}
\hline
New features & 5-shot & 10-shot  \\
  \hline
  Noise & 58.63 $\pm$ 0.59 \% & 67.89 $\pm$ 0.60\%  \\
  
  Mixup \cite{zhang2018mixup} (ICLR'17) & 66.34 $\pm$ 0.55 \% & 75.92 $\pm$ 0.58\%  \\
CutMix \cite{yun2019cutmix} (ICCV'19) & 69.23 $\pm$ 0.57 \% & 78.45 $\pm$ 0.53\%  \\

Hybrid (Ours) & \textbf{70.97} $\pm$ 0.57\% & \textbf{79.72} $\pm$ 0.54\% \\
\hline

\end{tabular}
\label{tab:ablnoise}
\vspace{-15pt}
\end{table}

Table \ref{tab:ablnoise} depicts the performance of FEAT with our proposed RNNP using random Gaussian noise in place of hybrid features. This ablation experiment involves randomly sampling noise from a Gaussian distribution as unlabeled features instead of generating hybrid features from the support examples for our method. The results indicate that using noise significantly hurts the performance for both 5-shot and 10-shot episodes.

\subsection{Effect of using CutMix and Mixup}\label{sec:mixup}
In our method, we generate hybrid features from the support image features. Since we have very limited support examples per class, we cannot use the data generation techniques such as generative adversarial networks and variational auto encoders, which require a huge amount of data for training. We can also use data augmentation techniques such as CutMix \cite{yun2019cutmix} or MixUp \cite{zhang2018mixup} to produce hybrid images for our method. We can use the features of these hybrid images as unlabeled features for our method.

Table \ref{tab:ablnoise} depicts the performance of FEAT with our proposed RNNP using CutMix \cite{yun2019cutmix} or MixUp \cite{zhang2018mixup} to produce hybrid images from the support images. We use the features extracted from these hybrid images in place of the hybrid features generated by our approach. The results indicate that when we use mixup in our method, the resulting performance is significantly lower than our method's performance using hybrid features. When we use CutMix in our method, the results are better than that of mixup but still lower than our method's performance with hybrid features. A possible reason for this is that both these methods are used for increasing the training data (data augmentation), and in this case, we do not train the base network $B$.

\subsection{Comparison with Noisy Label Approaches}\label{sec:jocor}
\begin{table}[t]
\footnotesize
\centering
\vspace{-5pt}
\caption{Performance of FEAT with RNNP/JoCoR/Co-Teaching+ on the mini-ImageNet dataset over 5/10-shot 5-way episodes with 40\% label corruption.}
\begin{tabular}{l|c|c}
\hline
Noise Correction & 5-shot & 10-shot  \\
  \hline
  Co-Teaching+ \cite{yu2019does} (ICML'19) & 59.34 $\pm$ 0.56 \% & 68.23 $\pm$ 0.57\%  \\

JoCoR \cite{Wei_2020} (CVPR'20) & 59.89 $\pm$ 0.55 \% & 68.56 $\pm$ 0.59\%  \\

RNNP (Ours) & \textbf{70.97} $\pm$ 0.57\% & \textbf{79.72} $\pm$ 0.54\% \\
\hline

\end{tabular}
\label{tab:abljocor}
\vspace{-15pt}
\end{table}

As discussed in Sec. \ref{sec:relnoise}, there are many approaches for learning from data with noisy labels. However, these methods require large amounts of data for training. In our setting, the novel classes, which contain the corrupted labels, have very few support examples. Therefore, these methods suffer from overfitting and do not perform well. To validate this point, we experimented with JoCoR \cite{Wei_2020} and Co-Teaching+ \cite{yu2019does} in this setting. For every novel class episode, we trained the base network $B$ on the support examples using JoCoR and Co-Teaching+. Both methods require two networks for training, and so we make two copies of the base network for each episode. We do not carry forward the knowledge from one episode to the next, as mandated during the novel class evaluation phase. Table \ref{tab:abljocor} reports the results for these experiments. The results indicate that the network cannot perform well using these methods because of the extremely limited training data.

\section{Analysis}
\subsection{Effect of Prototype Refinement}

\begin{figure}[t]
\includegraphics[width=0.47\textwidth]{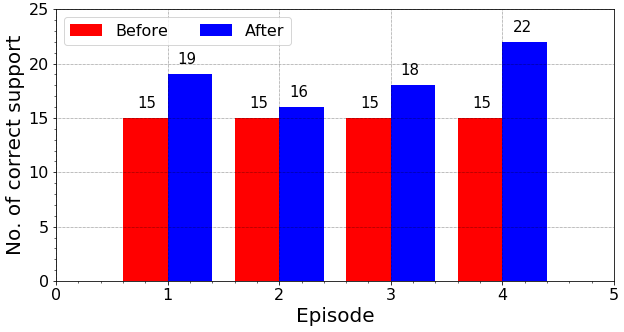}
\vspace{-10pt}
\caption{Effect of FEAT with RNNP on the support examples with noisy labels in the novel class 5-shot 5-way episodes of mini-ImageNet with 40\% label corruption.}
\label{fig:refineeffect}
\vspace{-5pt}
\end{figure}

We perform experiments to analyze how our method RNNP rectifies the incorrectly labeled support examples. Fig.~\ref{fig:refineeffect} depicts the change in the number of support examples that have correct labels after RNNP is applied (on FEAT backbone) for four randomly chosen 5-shot 5-way episodes with 40\% label corruption. The figure indicates that our method consistently corrects many incorrectly labeled support examples in each episode. Our method is able to rectify the labels of support examples and thereby refine the class prototypes. The refined prototypes help to achieve better performance in this setting.

\subsection{Statistical Significance}
\begin{figure}[t]
\subcaptionbox{}{
\includegraphics[width=0.225\textwidth]{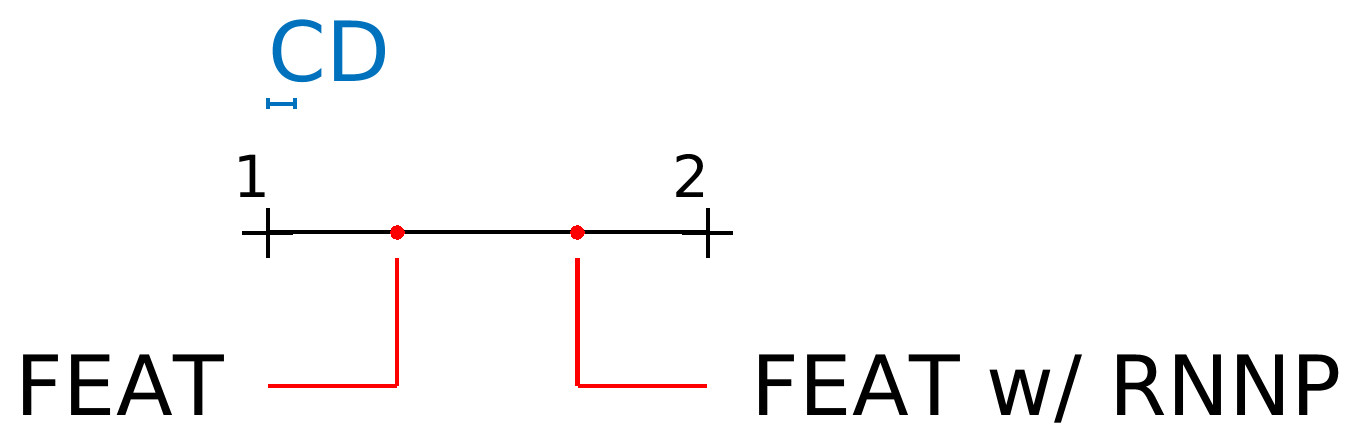}}
\subcaptionbox{}{
\includegraphics[width=0.23\textwidth]{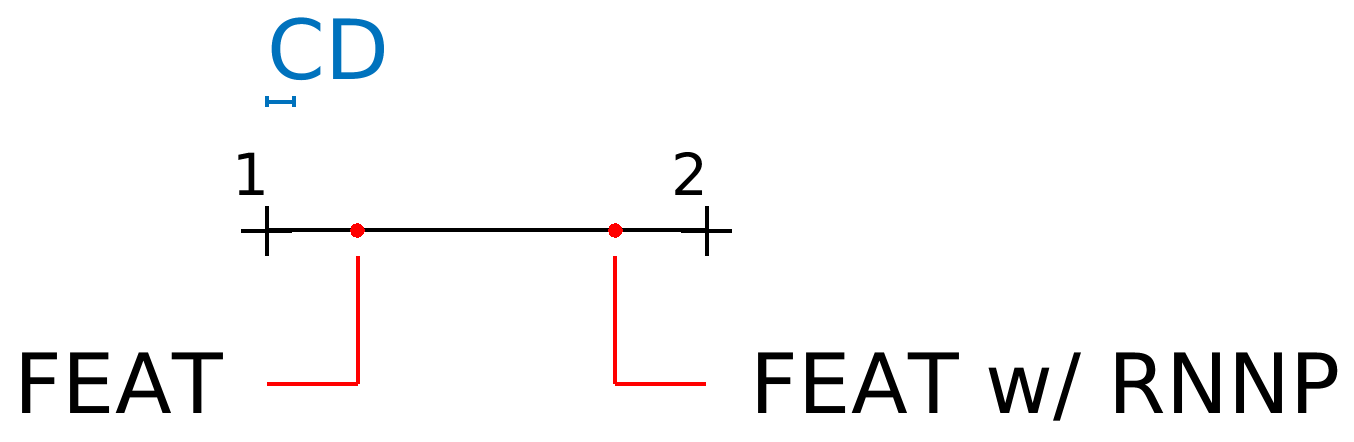}}
\vspace{-10pt}
\caption{Statistical significance test to show that FEAT w/ RNNP is statistically different from FEAT for 5-shot 5-way novel class episodes of mini-ImageNet dataset with (a) 20\% label corruption rate (b) 40\% label corruption rate with a significance level of 0.05.}
\label{fig:statsig}
\vspace{-15pt}
\end{figure}

We studied the statistical significance \cite{demvsar2006statistical} for FEAT with our method RNNP against FEAT. If the difference in the rank of the two methods lies within Critical Difference \cite{demvsar2006statistical}, then they are not significantly different. Fig.~\ref{fig:statsig} visualizes the post hoc analysis using the CD diagram for few-shot classification with noisy support labels. We observe in Fig.~\ref{fig:statsig} that the statistical difference between FEAT with our method RNNP (FEAT w/ RNNP) and FEAT is more than the Critical Difference (CD=0.062) for both 20\% and 40\% label corruption rates. Therefore, FEAT with RNNP is statistically different from FEAT.

\section{Conclusion}
We propose a novel approach called Robust Nearest Neighbor Prototype-based testing (RNNP) that makes the few-shot testing process robust to noisy labels in the novel class episodes. Our approach does not require any additional training and can replace the testing process of any few-shot learning method that uses a nearest neighbor prototype-based testing strategy. We empirically show that our method improves the performance of multiple few-shot learning methods in the presence of noisy labels. Through extensive validation experiments, we validate the different components of our method.

{\small
\bibliographystyle{ieee_fullname}
\bibliography{egbib}
}

\end{document}